\documentclass[runningheads]{llncs}
\usepackage{graphicx}
\usepackage{amsmath}
\usepackage{multirow}
\usepackage{booktabs}
\begin{document}
\title{Lighting Enhancement Aids Reconstruction of Colonoscopic Surfaces}
%
%
\author{Yubo Zhang\inst{1}\orcidID{0000-0002-8942-5253} \and
Shuxian Wang\inst{1}\orcidID{0000-0002-0184-3021} \and
Ruibin Ma\inst{1}\orcidID{0000-0002-1209-3588} \and
Sarah K. McGill\inst{2}\orcidID{0000-0002-4006-2703} \and
Julian G. Rosenman\inst{3}\orcidID{0000-0001-7500-0492} \and
Stephen M. Pizer\inst{1}\orcidID{0000-0002-4250-6531}
}
\authorrunning{Y. Zhang et al.}
%
\institute{Department of Computer Science, University of North Carolina at Chapel Hill \\
\email{\{zhangyb,shuxian,ruibinma,pizer\}@cs.unc.edu} \and
Department of Medicine, University of North Carolina at Chapel Hill \\
\email{mcgills@email.unc.edu } \and
Department of Radiation Oncology, University of North Carolina at Chapel Hill \\
\email{rosenmju@med.unc.edu}
}
\maketitle              
\begin{abstract}
High screening coverage during colonoscopy is crucial to effectively prevent colon cancer.
Previous work has allowed alerting the doctor to unsurveyed regions by reconstructing the 3D colonoscopic surface from colonoscopy videos in real-time.
However, the lighting inconsistency of colonoscopy videos can cause a key component of the colonoscopic reconstruction system, the SLAM optimization, to fail.
In this work we focus on the lighting problem in colonoscopy videos.
To successfully improve the lighting consistency of colonoscopy videos,
we have found necessary a lighting correction that adapts to the intensity distribution of recent video frames.
To achieve this in real-time, we have designed and trained an RNN network.
This network adapts the gamma value in a gamma-correction process. 
Applied in the colonoscopic surface reconstruction system, our light-weight model significantly boosts the reconstruction success rate, making a larger proportion of colonoscopy video segments reconstructable
and improving the reconstruction quality of the already reconstructed segments.

\keywords{Colonoscopy, image enhancement, 3D reconstruction}
\end{abstract}

\section{Introduction}
Colonoscopy is an effective examination to prevent colon (large intestine) cancer by screening for lesions.
During a colonoscopy a flexible tube called a colonoscope is inserted up to the distal end of the patient's colon, and then it is withdrawn through the colon while producing a video
by a camera that is attached at the tip of the colonoscope.
A point light attached to the colonoscope tip moves with the camera to provide the lighting source.
During a colonoscopy the camera sends back live images of the colon wall to the doctor,
who watches, detects and removes cancerous or pre-cancerous lesions with a surgical knife inside the colonoscope.

\begin{figure}
    \centering
    \includegraphics[width=0.8\textwidth]{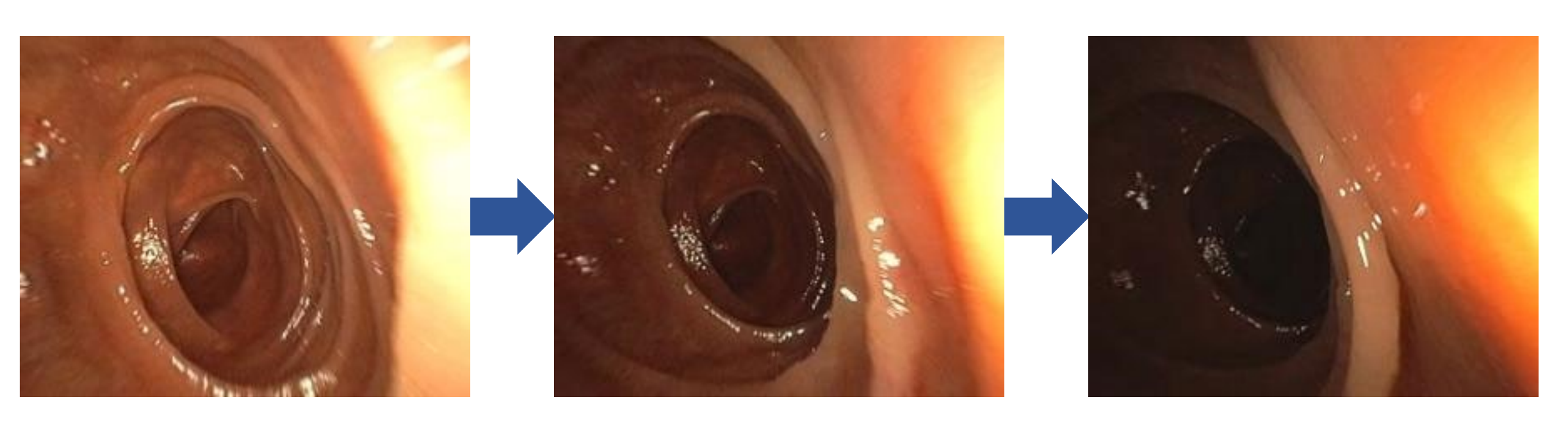}
    \caption{The lighting in the colonoscopy video can change rapidly from frame to frame, and a large part of the colon interior can be poorly lit due to the complicated colon terrain. This lighting issue can bring troubles to the diagnosis and the colon surface reconstruction system.}
    \label{fig:intro}
\end{figure}

Due to time limitations, colonoscopy is a one-pass exam, but high coverage of screening and high polyp detection rate are crucial.
Previous work tried to aid the colonoscopy process by alerting the doctor to unsurveyed colon regions (``blind spots'') revealed by reconstructing the already screened surfaces in real-time from colonoscopy videos~\cite{ma2019real}.
However, as shown in Fig.~\ref{fig:intro}, the complex topography makes a notable proportion of the colon surface be poorly lit;
also, it causes
brightness and contrast to change rapidly from frame to frame.
The poor lighting situation challenges 1) lesion detection by the doctor and 2) successful reconstruction of the surfaces, thereby allowing alerts as to blind spots.

In this work we focus on this lighting issue in colonoscopy.
We aim to make the lighting of colonoscopy videos consistent between consecutive frames and also to brighten the dark regions in colonoscopy video frames.
Such image enhancement is commonly carried out by curve adjustment approaches~\cite{huang2012efficient,guo2020zero} such as gamma correction, and recent deep learning methods can apply a more sophisticated adjustment to each image pixel with CNN- \cite{gharbi2017deep,wang2019underexposed} or GAN-based models~\cite{chen2018deep,jiang2019enlightengan}.
Considering that our inputs are video frames, we adopt the learning-based approach and develop an RNN-based network for adaptive curve adjustment.
Our network is trained in an unsupervised fashion using a loss measuring the lighting consistency among nearby frames and will
produce well-lit frame sequences at test time.
Applying our model as a pre-processing step to the frames for reconstruction leads to significant improvement of both the number of reconstructable colonoscopy video segments and the quality of reconstructions, with time overhead small enough to be ignored.

Overall, the contributions of this work are
\vspace{-4pt}
\begin{itemize}
  \item To our knowledge, our work is the first one to focus on the lighting consistency issue in the colonoscopy video.
  \item We propose a light-weight RNN model for lighting enhancement of an image sequence, which can be trained without ground-truth. Tested on real colonoscopy images, our model can effectively brighten the poorly lit regions in each frame, and make the lighting consistent from frame to frame.
  \item Most importantly, applied in the colonoscopic surface reconstruction system, our model effectively boosts the reconstruction success rate and improves the reconstruction quality without sacrificing the time efficiency of the system.
\end{itemize}
\section{Background: SLAM and colonoscopic reconstruction}
Here, we review the mechanism behind the 3D reconstruction technique SLAM, which is a key component in colonoscopic reconstruction systems.
We then analyze why the lighting issue in colonoscopy can cause the SLAM system to fail.

\subsection{SLAM mechanism}
3D reconstruction is a challenging task in computer vision.
One of the most successful methods for this task is Simultaneous Localization And Mapping (SLAM).
SLAM systems have achieved tremendous success in indoor and outdoor-scene reconstruction~\cite{geiger2013vision,sturm12iros}.
Recently, with the development of deep learning, neural networks have been applied to aid SLAM for better reconstruction~\cite{tateno2017cnn,wang2017deepvo,yang2018deep}, especially in the more challenging scenarios such as reconstruction from colonoscopy video~\cite{ma2019real}.

SLAM is an algorithm that can achieve real-time dense reconstruction from a sequence of monocular images~\cite{newcombe2011dtam,engel2013semi,mur2015orb,engel2017direct}.
As the name suggests, SLAM has a localization component and a mapping component; the two components operate cooperatively.
The localization (tracking) component predicts the camera poses from each incoming image frame. Based on the visual clues extracted from the images, the mapping component optimizes especially the pose predictions but also the keypoints' depth estimates.
The objective function used in the optimization of the SLAM of~\cite{engel2017direct}, which is applied in colonoscopic reconstruction~\cite{ma2019real}, is
\begin{gather} \label{Eq-SLAM}
    E_{ij} = \sum_{\boldsymbol{p}\in \mathcal{P}_j} \omega_{\boldsymbol{p}}	\left\lVert ({I_{j}[\boldsymbol{p}']-b_j)-\frac{e^{a_j}}{e^{a_i}}(I_i[\boldsymbol{p}]-b_i)} \right\rVert_{\gamma} \\
    \boldsymbol{p}' = {\rm \Pi} (T {\rm \Pi}^{-1} (\boldsymbol{p},d_{\boldsymbol{p}}))
\end{gather}
Sampled from a source image $I_i$, each keypoint $\boldsymbol{p}$ in the keypoint set $\mathcal{P}_j$ can be projected to a location $\boldsymbol{p}'$ in a target image $I_j$, based on its predicted depth $d_{\boldsymbol{p}}$ and the predicted camera transformation $T$.
${\rm \Pi}$ denotes the projection.
Supposing the transformation and depth predictions are correct, $\boldsymbol{p}$ of image $I_i$ and $\boldsymbol{p}'$ of image $I_j$ come from the same surface point, so $I_i[\boldsymbol{p}]$ and $I_j[\boldsymbol{p'}]$ should have the same intensity.
This assumption is the mechanism behind the energy function $E_{ij}$ which is the photometric error between two frames $I_i$ and $I_j$ in Eq.~\ref{Eq-SLAM},
where $\left\lVert * \right\rVert_{\gamma}$ denotes the Huber norm.
The $a_k$ and $b_k$ values are discussed below.

\subsection{The lighting problem in colonoscopic surface reconstruction}
The photometric-based optimization in SLAM depends on the light consistency between frames, that assuming the same physical location in the environment is shown in similar intensities in different frames.
This is a prerequisite that can be mostly fulfilled in an indoor or outdoor scenario with the steady lighting source, e.g., sunlight.
When it comes to colonoscopy, this lighting consistency assumption can often be violated since the point light is moving with the camera and can change rapidly due to motion and occlusion.
Although the SLAM algorithm also optimizes an additional brightness transformation as a compensation, denoted as $e^{-a}(I-b)$ in Eq.~\ref{Eq-SLAM}, it just fixes each frame's exposure and cannot handle the more complicated lighting changes in colonoscopy videos such as contrast difference (bright regions become brighter and dark regions become darker).
So without specific brightness adjustment, this kind of rapid lighting change in colonoscopy makes the SLAM system unstable, often leading to tracking failure.

We want to alleviate the failures caused by lighting changes.
In this work we accomplished this by explicitly adjusting the brightness of the image sequence to make them more consistent.
This requires solving an adaptive image enhancement problem that each image in the sequence is enhanced in relation to adjacent frames.
We develop a deep learning method for this task, which will be discussed in detail in the next section.

\section{Method}
In this work we apply an adaptive intensity mapping to enhance the colonoscopy frame sequence with the help of an RNN network,
whose implementation and unsupervised training strategy will be introduced in this section.

\subsection{Adaptive gamma correction}
Image enhancement is a classical topic in computer vision.
Multiple directions have been proposed to resolve the issue, such as histogram equalization and its variants~\cite{pizer1987adaptive,xiao2019histogram}, unsharp masking~\cite{polesel2000image,luft2006image} and more recently deep learning pixel-wise prediction approaches~\cite{gharbi2017deep,jiang2019enlightengan,ke2020gct}.

\begin{figure}[ht]
    \centering
    \includegraphics[width=1.0\textwidth]{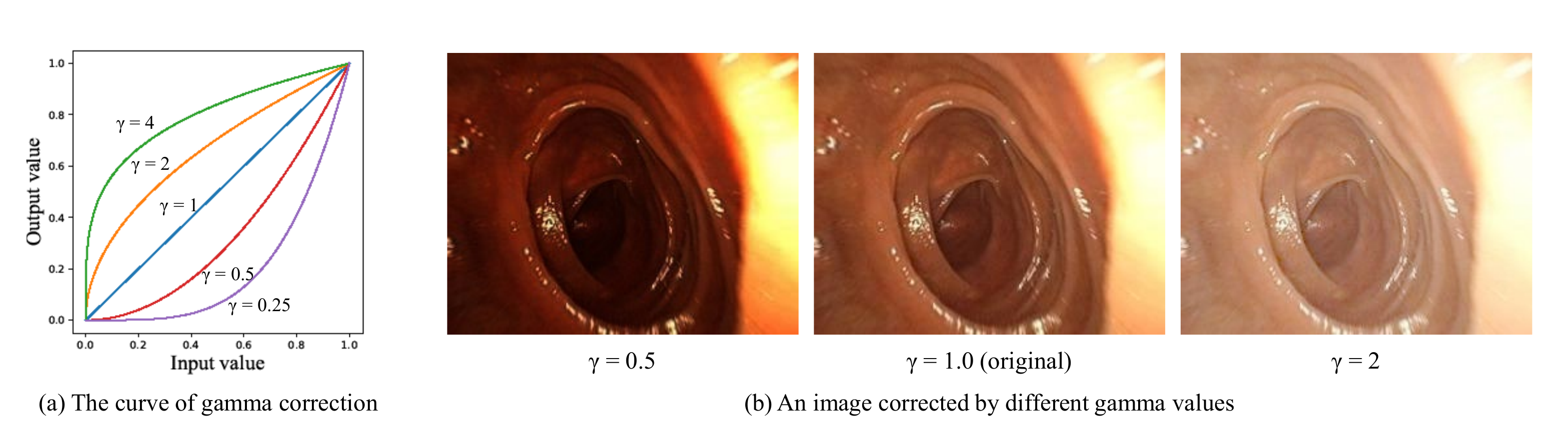}
    \caption{The gamma correction: (a) Gamma correction's effect on image intensity; \\
    (b) The figure in the middle is been adjusted by different gamma values.}
    \label{fig:gamma}
\end{figure}

Our driving problem of the reconstruction from colonoscopy video requires real-time execution,
so our lighting enhancement must be accomplished at the video frame rate.
This requires that only a handful of parameters be adjusted from frame to frame in an enhancement method.
An example approach that meets this requirement is gamma correction~\cite{huang2012efficient,guo2020zero}.
In gamma correction, for every pixel in an image, its input value $I_{in}$ is adjusted by the power $\gamma$, then multiplied by a constant $A$ to get the enhanced value $I_{out}$:
\begin{gather}
    I_{out} = A I_{in}^\gamma
\end{gather}
In practice, the input intensities will be normalized to $[0,1]$, so the constant $A$ will be set to $1$ and $\gamma$ is the only parameter controlling the adjustment.
By applying a different $\gamma$ value, the same image can be brightened or darkened to a different extent,
as the example shown in (b) of Fig.~\ref{fig:gamma}.

\begin{figure}[ht]
    \centering
    \includegraphics[width=1.0\textwidth]{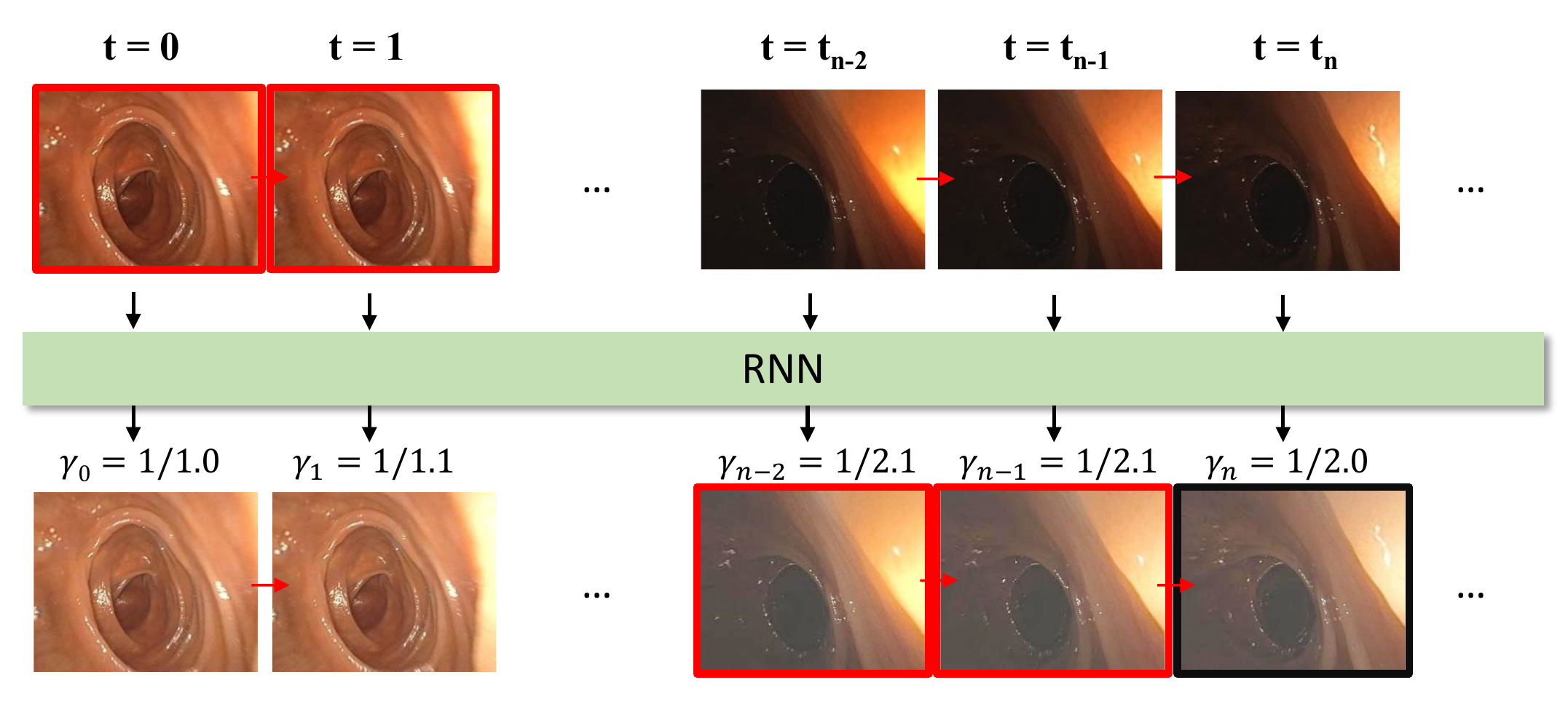}
    \caption{At each time step, the RNN takes an input image and predicts its $\gamma$ value to enhance the image. The network is trained by contrasting the current corrected image with two previous corrected images and the first two input images of the sequence.}
    \label{fig:network}
\end{figure}

Our goal is to improve the lighting consistency of an image sequence.
Adjusting the way in which each frame is enhanced requires an adaptive version.
To model this adaptation, we specifically design a recurrent neural network (RNN) to capture the temporal information needed in the enhancement and to predict the $\gamma$ value for each image.
The overall pipeline of our RNN-based adaptive gamma correction is shown in Fig.~\ref{fig:network}.
The details of the RNN network will be discussed next.

\subsection{RNN network}
Although the gamma correction is cheap in computation,
our enhancement RNN network also needs to be light-weight in order to serve as an additional pre-processing step for the current real-time colonoscopic surface reconstruction system.
Thus, we build our RNN network on the ConvLSTM unit~\cite{xingjian2015convolutional}, with only two additional convolution layers:
\begin{gather}
    x_1 = {\rm ReLU}({\rm Conv}(I_i)) \\
    [x_2, h_i] = {\rm ConvLSTM}([x_1, h_{i-1}]) \\
    x_3 = {\rm ReLU}({\rm Conv}(x_2)) \\
    f = {\rm AvgPool}(x_3) \\
    \gamma_i = {\rm ReLU}(Af  + b)
\end{gather}
At test time, at each time step the network takes the current image frame $I_i$ as the input, includes latent information of the previous frames $h_{i-1}$ into the computation, and predicts the gamma correction value $\gamma_i$ for the current frame.
The prediction happens recurrently, enhancing the entire sequence as shown in Fig.~\ref{fig:network}.

\subsection{Training strategy}
With no ground-truth to supervise the network training, our RNN is trained in an unsupervised fashion to achieve lighting consistency.
We achieve this consistency by comparing the current adjusted frame to four reference frames.
Only in training, these reference frames consist of the previous two adjusted frames and the first two images of the entire sequence, which serve as the ``seed'' images to stabilize the training by setting a brightness baseline.
For example in Fig.~\ref{fig:network}, the target frame is surrounded in black and all the reference frames are surrounded in red.

The loss we are optimizing utilizes the structural similarity measurement~\cite{wang2004image}:
\begin{align}
    L_{ssim} &= {\rm mean}(1 - {\rm SSIM}(I_r, I_t)) \\
    &= {\rm mean} (1 - \frac{(2\mu_{I_r}\mu_{I_t} + c_1) (2\sigma_{I_rI_t} + c_2)} {(\mu_{I_r}^2 + \mu_{I_t}^2 + c_1) (\sigma_{I_r}^2 + \sigma_{I_t}^2 + c_2)})
\end{align}
where at each pixel location, $\mu_{I}$ is the local average, $\sigma_{I}^2$ is the local variance and $\sigma_{I_rI_t}$ is the covariance of two images.
$c_1$ and $c_2$ are small constants to avoid dividing by zero.
The SSIM function is composed of three comparisons: luminance, contrast and structure.
When applying to training, the luminance and contrast comparisons will force the network to produce adjusted images with similar lighting to the reference frames.

Specular regions have very high intensities that inordinately influence the lighting estimation in a frame.
Therefore, in training when computing the $L_{ssim}$, we mask out the pixels with input intensity larger than $0.7$. 
Each training sequence contains $10$ sequential frames, with the first two as seeds, and the loss of each target frame is computed as the average $L_{ssim}$ of its four reference frames.
\section{Experiments}
\subsection{Implementation details}
To build the dataset of our task, we collected $105$ colonoscopy video snippets, each containing $50$ to $150$ frames.
$60$ snippets were used as the training data, $1$ was used for validation, and the rest were reserved for evaluation.
Among the $44$ evaluation sequences, $12$ of them could be successfully reconstructed using the colonoscopic surface reconstruction system in previous work~\cite{ma2019real}, while $32$ of them could not.

For the training and validation snippets, we divided them into $10$-frame overlapping sub-sequences.
In this way we created about $1800$ training sequences.
We used the Adam optimizer with a fixed learning rate of $5 \times 10^{-5}$ to train the RNN network for $10$ epochs.
A batch size of $4$ was used.
We implemented our method using the PyTorch framework on a single Nvidia Quadro RTX$5000$ GPU.

At test time our network enhances the entire snippet from start to end
with an average image enhancement run time of less than a millisecond.
Considering the frame rate we use to extract images from video is $18$ frames per second, the overhead brought by our method is small enough and does not violate the real-time execution when added to the colonoscopic reconstruction system.

\subsection{Visual effect}
\begin{figure}[ht]
\vspace{-5pt}
    \centering
    \includegraphics[width=1.0\textwidth]{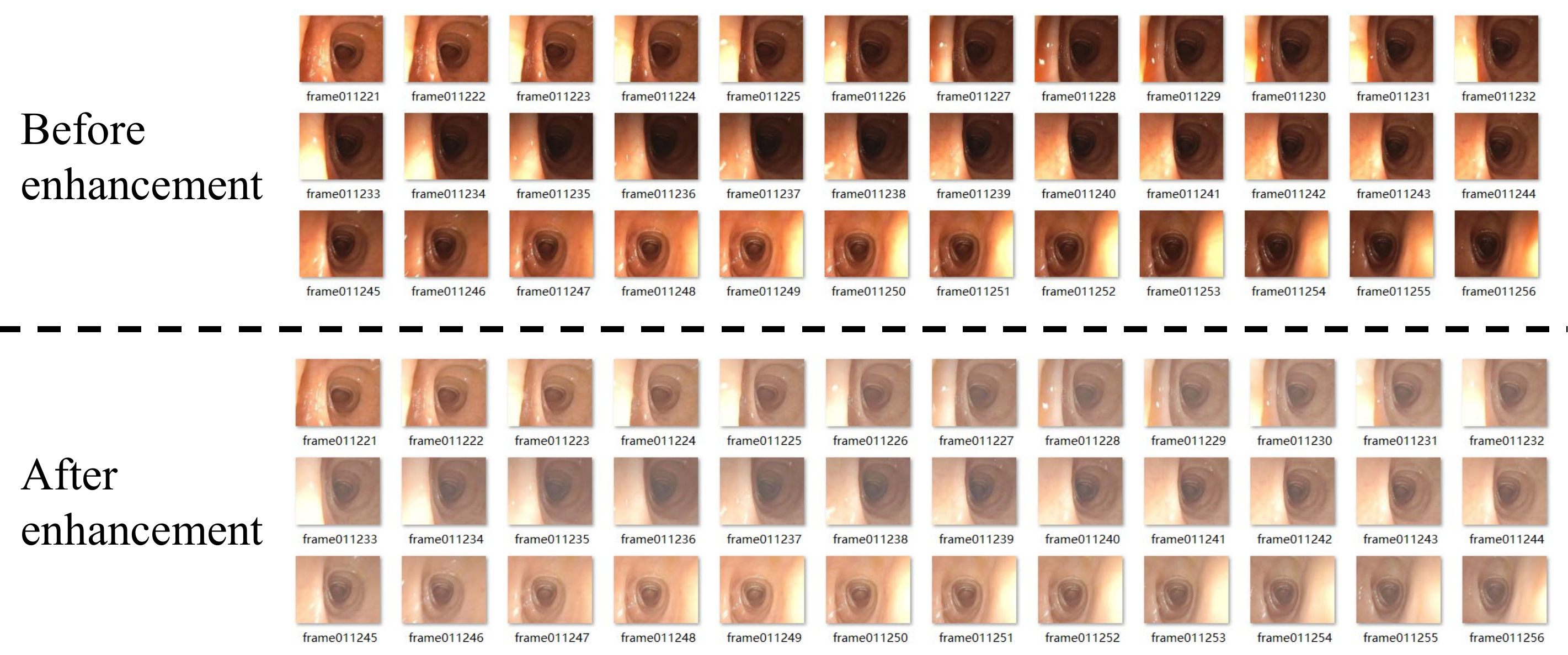}
    \caption{The visual comparison of the same sequence before (top) and after (bottom) the RNN enhancement. After enhancement, the extreme dark regions have been brightened up and the lighting consistency improves.}
    \label{fig:visual}
\vspace{-5pt}
\end{figure}

In Fig.~\ref{fig:visual} we show a snippet that is been enhanced using our trained RNN network, where the original frames are shown on the top and the enhanced ones on the bottom.
The overall brightness of the sequence has been largely improved and the ``down the barrel'' regions are brightened up and clearly revealed after enhancement.
Moreover, the lighting consistency is significantly improved.
The image contrast changes rapidly in the original sequence, due to a haustral ridge blocking much of the light.
After enhancement, these changes across frames from bright on average to dark on average and then back to bright on average are far less obvious: the lighting of the sequence becomes much more consistent.

\subsection{Application in colonoscopic surface reconstruction}
In the colonoscopic surface reconstruction system RNNSLAM \cite{ma2019real}, a deep learning RNN network runs in parallel with a standard SLAM system.
The RNN component predicts the depths and pose of each camera frame to initialize their SLAM optimization, and the SLAM outputs the improved values of poses and depths and updates the hidden states of RNN.
This combination leads to low tracking error.
Our RNN lighting enhancement network can also run in parallel with the RNNSLAM.
The enhanced images produced by our method are given to its SLAM component as the image inputs.

\subsubsection{Success on the cases that fail with original images}
To prove the validity of our method,
we tested the RNNSLAM system with enhanced images on $32$ cases which cannot be reconstructed with the original images.
A case is categorized as a failure when RNNSLAM predicts obvious discontinuous camera poses, or the system aborts automatically when it cannot give reasonable pose prediction.
We noticed that most of the previous failure cases include significant lighting changes or large camera motion that leads to lighting occlusion.
When testing them with our enhancement module inserted, $21$ of $32$ succeeded without tracking failure:
the reconstruction success rate significantly increased.
We also tested these $32$ cases with the images enhanced by histogram equalization~\cite{pizer1987adaptive} and by a neural-network method Zero-DCE~\cite{guo2020zero}\footnote[1]{We implemented the training in \cite{guo2020zero} without color constancy loss using the colonoscopy images from our training set.};
only $8$ and $12$, respectively succeeded, showing our method outperforms the traditional and deep-learning single-image enhancement approaches on improving the reconstruction success rate.

\begin{figure}[ht]
    \centering
    \includegraphics[width=1.0\textwidth]{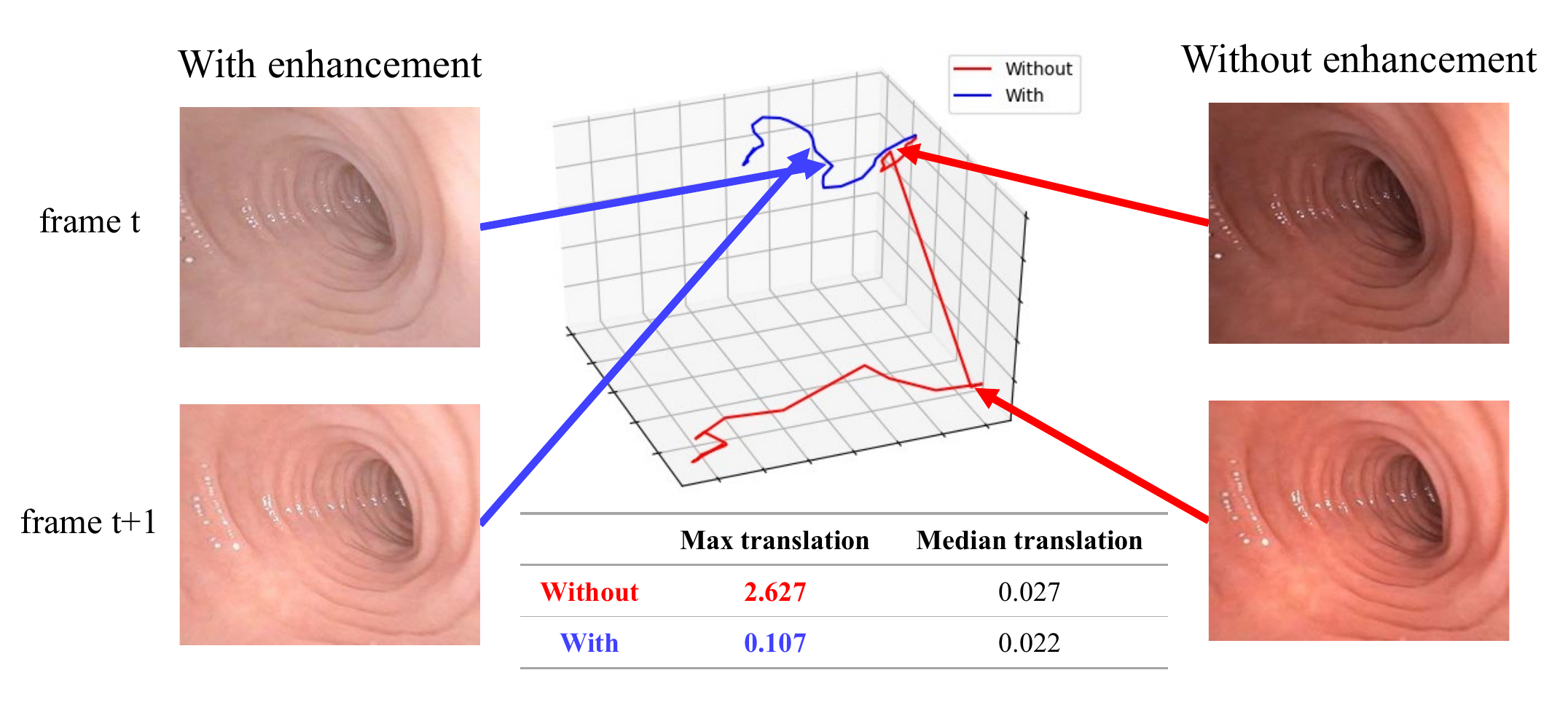}
    \caption{The comparison of pose trajectories for the same sequence without (red) and with (blue) the RNN lighting enhancement. The pose discontinuity issue indicated by large camera translation does not occur after using enhanced images.}
    \vspace{-5pt}
    \label{fig:traj}
    \vspace{-5pt}
\end{figure}

In Fig.~\ref{fig:traj} we illustrate a case of improved tracking using our lighting correction,
where the predicted camera trajectory without enhancement is shown in red, and the trajectory after enhancement is shown in blue.
In each case the camera is predicted to start from the top right position and to move along the respective trajectory to the bottom left position.
Particularly in the figure, we show the positions predicted for two successive keyframes $t$ and $t+1$.
Although it is clear to the human eye that the motion between these two images is subtle,
due to a lighting change, the camera pose prediction changes by a dramatic distance of $2.63$ when using the original frames, clearly indicating an error comparing to the $0.03$ median camera translation of this trajectory.
This issue of discontinuity in predicted poses does not occur after we enhance the input images, where the predicted camera poses move smoothly and the maximum camera translation of this trajectory is $0.11$.

\subsubsection{Pose improvement of previously reconstructable sequences}
Not only does our lighting enhancement allow previously un-reconstructable sequences to be reconstructed, but as demonstrated in the following, it improves the pose trajectories produced from sequences that were previously reconstructable.
To quantitatively measure the pose improvement, we adopt the evaluation method in~\cite{ma2019real}, using COLMAP~\cite{schoenberger2016sfm} on the un-enhanced images to generate a high-quality baseline trajectory as the virtual ``ground-truth" for each video sequence.
As the evaluation metric we compute the absolute pose error (APE) of each timestamp in the generated trajectory compared to its ``ground-truth".
For each trajectory, the RMSE of the APE statistics of all timestamps is computed.
We tested $12$ colonoscopic sequences that were already reconstructable without our enhancement strategy.
We computed the RMSE and other APE statistics (mean, std, etc.) of each sequence respectively, 
and in Table~\ref{table:ape} we list the summary statistics across these $12$ sequences.
For better reference, we also show the errors of the trajectories produced by COLMAP when using enhanced images.
Since we are measuring the error, the lower the result is, the better.

\begin{table}[ht]
\vspace{-10pt}
\begin{center}
\begin{tabular}{lcccccc}
\multirow{2}{*}{\bf Method} & mean & mean & std & min & median & max \\
 & RMSE & APE & APE & APE & APE & APE \\
\midrule
\midrule 
COLMAP w/ enhanced img & 
    0.329 & 0.295 & 0.143 & 0.077 & 0.278 & 0.736 \\
\midrule 
RNNSLAM w/ original img & 
    0.840  & 0.752 & 0.371 & 0.203 & 0.694 & 1.666 \\
RNNSLAM w/ enhanced img & \textbf{0.680} & \textbf{0.606} & \textbf{0.302} & \textbf{0.105} & \textbf{0.554} & \textbf{1.388} \\
\end{tabular}
\vspace{8pt}
\caption{The average APE statistics of 12 colonoscopic sequences}
\label{table:ape}
\vspace{-20pt}
\end{center}
\end{table}

For every sample in these $12$ sequences, even though the original lighting is consistent enough for the SLAM to produce decent pose predictions, our enhancement method brought further improvement, decreasing their pose errors.
This result shows that besides significantly increasing the reconstruction success rate, our method can also bring minor improvement to the reconstruction quality in cases that can be reconstructed successfully without lighting enhancement.

\section{Discussion and conclusion}
Due to the complexity of colon geometry, the lighting in colonoscopy videos tends to have rapid changes that lead to failures in the colonoscopic surface reconstruction system.
In this work we focused on improving the lighting consistency in colonoscopy videos.
We proposed a light-weight RNN network for an adaptive lighting enhancement method to enhance the colonoscopy image sequence, which brightens the dark regions and makes lighting consistent from frame to frame.
With the help of our enhancement module, a larger portion of colonoscopy videos can now be successfully reconstructed and the reconstruction quality is improved.

\subsubsection{Future work}
\begin{enumerate}
\item Our method is initially designed to improve the lighting consistency of colonoscopy sequences, but it is also applicable for other video modalities.
For example, for other types of endoscopy videos and outdoor driving sequences with large lighting changes, our network can be trained on these data to improve their lighting consistency for better reconstruction.
\item 
\begin{figure}[ht]
    \centering
    \includegraphics[width=1.0\textwidth]{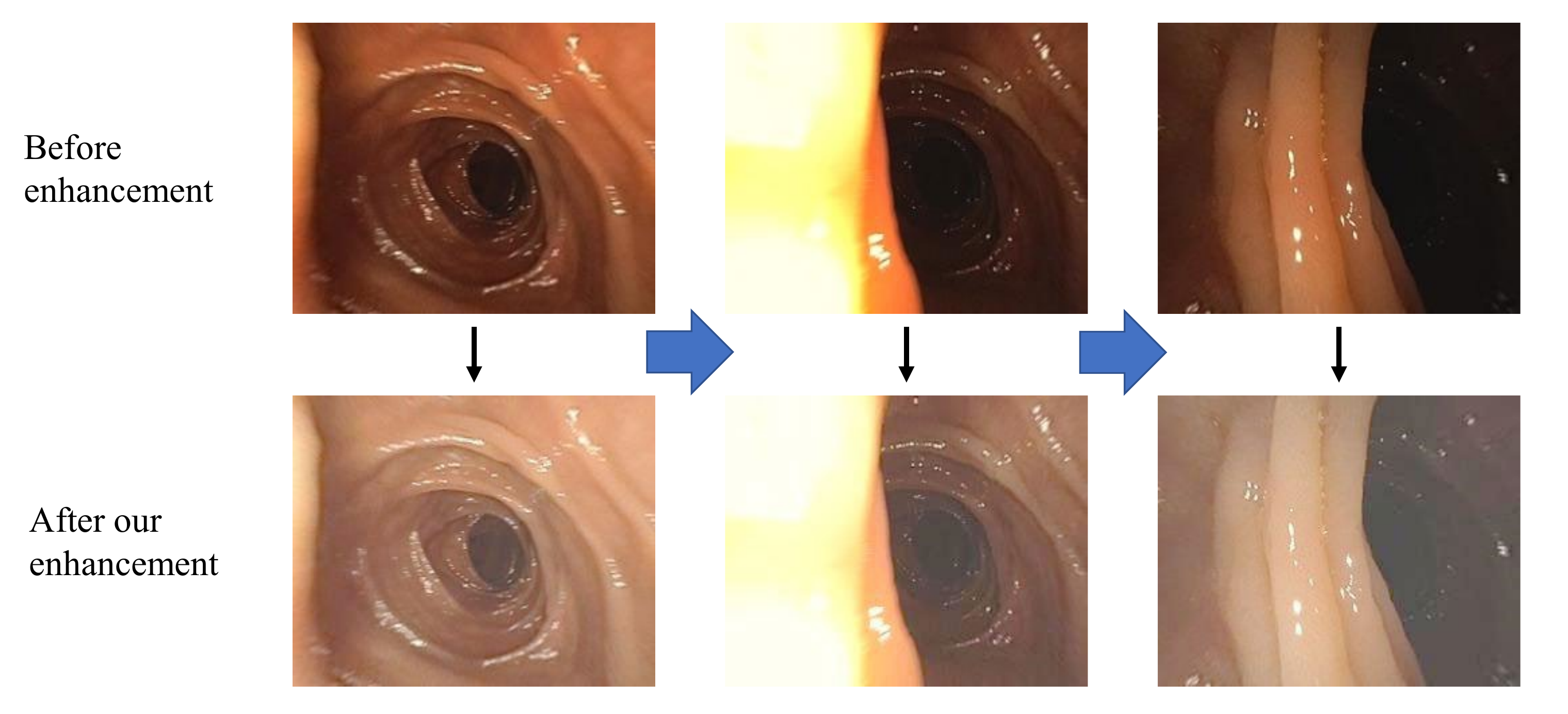}
    \caption{An extreme tracking condition, the ``close occluder''. Its complicated lighting situation is beyond our enhancement method's capacity.}
    \label{fig:close}
\end{figure}
Although our method successfully increases the reconstruction success rate, there is still a notable fraction of colonoscopy sequences that cannot be reconstructed after the lighting fix.
They usually contain some extreme tracking conditions,
and one of them is what we call the ``close occluder''.
In this case, as the colonoscope moving behind a haustral ridge, or moving side-way and being really close to the colon surface, the image foreground occupies a large portion of an image frame, as shown in the last two images of Fig.~\ref{fig:close}.
The lighting in these images is usually extremely bright in the foreground and extremely dark in the background.
Moreover, when using these frames, the SLAM system will choose keypoints for optimization not only from the ``down the barrel'' portion of the image, as it usually does with the normal frames, but also from the bright foreground.
Therefore, in order for these cases to succeed, the bright and the dark regions both need to be more elaborately adjusted to achieve global lighting consistency.
Currently our method cannot handle this complicated scenario and a more sophisticated image enhancement technique is needed;
we leave it to future work.

\end{enumerate}

\subsubsection{Acknowledgements.} We thank Prof. Jan-Michael Frahm for useful consultations. This work was carried out with financial support from the Olympus Corporation, the UNC Kenan Professorship Fund, and the UNC Lineberger Cancer Center.

%
%
%
\bibliographystyle{splncs04}
\bibliography{ref}
\end{document}